\documentclass[12pt]{article}

\usepackage{times}

\newenvironment{sciabstract}{%
\begin{quote} \bf}
{\end{quote}}
\usepackage[bookmarks=true]{hyperref}
\usepackage{pdfsync}
\usepackage[dvipsnames,table]{xcolor}
\usepackage{mathtools}
\usepackage{amsmath} 
\usepackage{amssymb}  
\usepackage{subcaption}
\usepackage{multirow}
\usepackage{float}
\usepackage{amsmath}
\usepackage{nccmath} 
\usepackage{algorithm}
\usepackage{pbox}
\usepackage[normalem]{ulem}

\newcommand{\separator}{ \noindent \rule{\columnwidth}{1pt} }

\newcommand{\doubleBlind}[1]{} 

\newcommand{\lt}{{\tt True}}
\newcommand{\lf}{{\tt False}}

\newfloat{spec}{thb}{lop} 
\floatname{spec}{Specification}
\makeatletter
\newcommand{\leqnomode}{\tagsleft@true}
\newcommand{\reqnomode}{\tagsleft@false}
\makeatother
\title{An Integrated System for Perception-Driven Autonomy with Modular Robots}

\author
{Jonathan Daudelin\textsuperscript{*}, Gangyuan Jing\textsuperscript{*}, Tarik Tosun\textsuperscript{*}, \\ Mark Yim, Hadas Kress-Gazit, and Mark Campbell\\
\\
\normalsize{Daudelin, Jing, Kress-Gazit, and Campbell: Department of Mechanical and Aerospace Engineering,}\\
\normalsize{Cornell University, Ithaca NY, USA}\\
\normalsize{Tosun and Yim: Department of Mechanical Engineering and Applied Mechanics, }\\
\normalsize{University of Pennsylvania, Philadelphia PA, USA}\\
\\
\normalsize{\textsuperscript{*} J. Daudelin, G. Jing, and T. Tosun contributed equally to this work.} \\
\normalsize{Address correspondence to: Jonathan Daudelin, jd746@cornell.edu} \\
}

\date{}

\usepackage{graphicx}
\usepackage{makeidx}
\makeindex

\begin{document}

\maketitle 

%
\begin{sciabstract}
The theoretical ability of modular robots to reconfigure in response to complex tasks in \textit{a priori} unknown environments has frequently been cited as an advantage and remains a major motivator for work in the field.

We present a modular robot system capable of autonomously completing high-level tasks by reactively reconfiguring to meet the needs of a perceived, \textit{a priori} unknown environment.  The system integrates perception, high-level planning, and modular hardware, and is validated in three hardware demonstrations. Given a high-level task specification, a modular robot autonomously explores an unknown environment, decides when and how to reconfigure, and manipulates objects to complete its task.

The system architecture balances distributed mechanical elements with centralized perception, planning, and control.  By providing an example of how a modular robot system can be designed to leverage reactive reconfigurability in unknown environments, we have begun to lay the groundwork for modular self-reconfigurable robots to address tasks in the real world.
\end{sciabstract}


%
\section{Introduction}
\label{sec:introduction}
Modular self-reconfigurable robot (MSRR) systems are composed of repeated robot elements (called \textit{modules}) that connect together to form larger robotic structures, and can \textit{self-reconfigure}, changing the connective arrangement of their own modules to form different structures with different capabilities.  Since the field was in its nascence, researchers have presented a vision that promised flexible, reactive systems capable of operating in unknown environments. MSRRs would be able to enter unknown environments, assess their surroundings, and self-reconfigure to take on a form suitable to the task and environment at hand \cite{Yim1994}.  Today, this vision remains a major motivator for work in the field \cite{Yim2007a}.  

Continued research in  MSRR has resulted in substantial advancement.  Existing research has demonstrated MSRR self-reconfiguring, assuming interesting morphologies, and exhibiting various forms of locomotion, as well as methods for programming, controlling, and simulating modular robots \cite{Yim1994, Yim2007, Rubenstein2004,Murata2006,Paulos2015,Jing2016, fukuda1990cellular, murata1994self, chirikjian1994kinematics, dutta2018distributed,ryland2010design, wolfe2012m, romanishin20153d, mantzouratos2015embeddability}.
However, achieving autonomous operation of a self-reconfigurable robot in unknown environments requires a system with the ability to explore, gather information about the environment, consider the requirements of a high-level task, select configurations whose capabilities match the requirements of task and environment, transform, and perform actions (such as manipulating objects) to complete tasks.  Existing systems provide partial sets of these capabilities.
Many systems have demonstrated limited autonomy, relying on beacons for mapping  \cite{Grabowski2000,Dorigo2005} and human input for high-level decision making \cite{Mondada2005,Dorigo2013}. Others have demonstrated swarm self-assembly to address basic tasks such as hill-climbing and gap-crossing \cite{gross2006autonomous,o2010self}.  While these existing systems all represent advancements, none have demonstrated fully autonomous, reactive self-reconfiguration to address high-level tasks.

This paper presents a  system allowing modular robots to complete complex high-level tasks autonomously.  The system automatically selects appropriate behaviors to meet the requirements of the task and constraints of the perceived environment.  Whenever the task and environment require a particular capability, the robot autonomously self-reconfigures to a configuration that has that capability.
The success of this system is a product of our choice of system architecture, which balances distributed and centralized elements.  Distributed, homogeneous robot modules provide flexibility, reconfiguring between morphologies to access a range of functionality.  Centralized sensing, perception, and high-level mission planning components provide autonomy and decision-making capabilities.  Tight integration between the distributed low-level and centralized high-level elements allows us to leverage advantages of distributed and centralized architectures.

The system is validated in three hardware demonstrations, showing that,  given a high-level task specification, the modular robot autonomously explores an unknown environment, decides if, when, and how to reconfigure, and manipulates objects to complete its task. By providing a clear example of how a modular robot system can be designed to leverage reactive reconfigurability in unknown environments, we have begun to lay the groundwork for reconfigurable systems to address tasks in the real world. 

\section{Results}
\label{sec:results}
%
We demonstrate an autonomous, perception-informed, modular robot system that can reactively adapt to unknown environments via reconfiguration to perform complex tasks. The system hardware consists of a set of \textbf{robot modules} (that can move independently and dock with each other to form larger morphologies), and a \textbf{sensor module} that contains multiple cameras and a small computer for collecting and processing data from the environment. Software components consist of a \textbf{high-level planner} to direct robot actions and reconfiguration, and \textbf{perception algorithms} to perform mapping, navigation, and classification of the environment.
Our implementation is built around the SMORES-EP modular robot \cite{tosun2016design}, but could be adapted to work with other modular robots.

Our system demonstrated high-level decision-making in conjunction with reconfiguration in an autonomous setting.  In three hardware demonstrations, the robot explored an \textit{a priori} unknown environment, and acts autonomously to complete a complex task.  Tasks are specified at a high level: users do not explicitly specify which configurations and behaviors the robot should use; rather, tasks are specified in terms of \textit{behavior properties}, which describe desired effects and outcomes \cite{JingAURO2017}.  During task execution, the high-level planner gathers information about the environment and reactively selects appropriate behaviors from a design library, fulfilling the requirements of the task while respecting the constraints of the environment.  Different configurations of the robot have different capabilities (sets of behaviors).  Whenever the high-level planner recognizes that task and environment require a behavior the current robot configuration cannot execute, it directs the robot to reconfigure to a different configuration that can execute the behavior.
        
\begin{figure}[H]
  \begin{subfigure}{0.5\columnwidth}
  \begin{center}
  \includegraphics[width=\columnwidth]{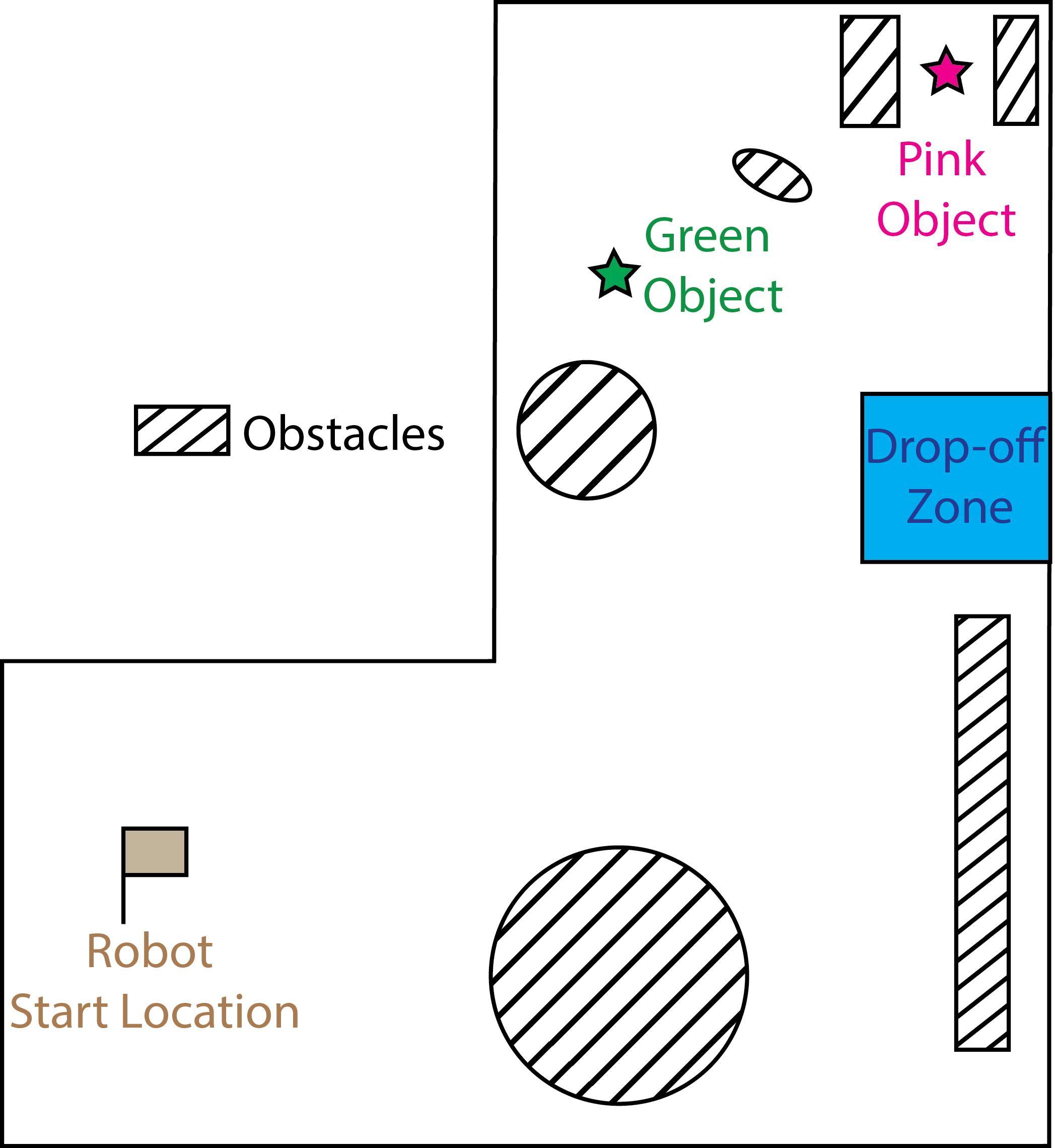}
  \caption{Diagram of Demonstration I environment}
  \label{fig:map}
  \end{center}
  \end{subfigure}
  %
  \begin{subfigure}{0.5\columnwidth}
  \begin{center}
  \includegraphics[width=\columnwidth]{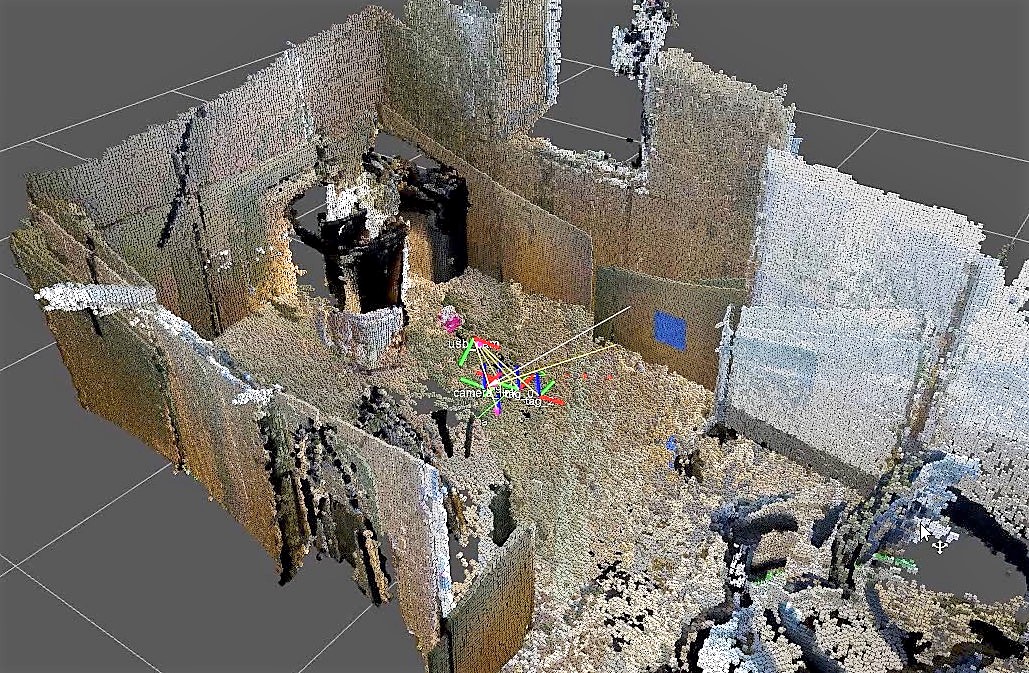}
  \caption{Map of environment 1 built by visual SLAM}
  \label{fig:octomap}
  \end{center}
  \end{subfigure}
  %
  \newcommand{\Lwidth}{0.4\columnwidth}
  \newcommand{\Rwidth}{0.4\columnwidth}
  \newcommand{\Rboxheight}{-0.5\height}
  \setlength{\tabcolsep}{4pt} 
  \begin{subfigure}{\columnwidth}
  \centering
  \begin{tabular}{|c|c|}
  \hline
   & \vspace{-5pt}\\

  \textbf{Environment Setup} & \textbf{Task Description}\\

  \hline
   & \vspace{-5pt}\\
   
  \raisebox{\Rboxheight}{\includegraphics[width=0.3\columnwidth]{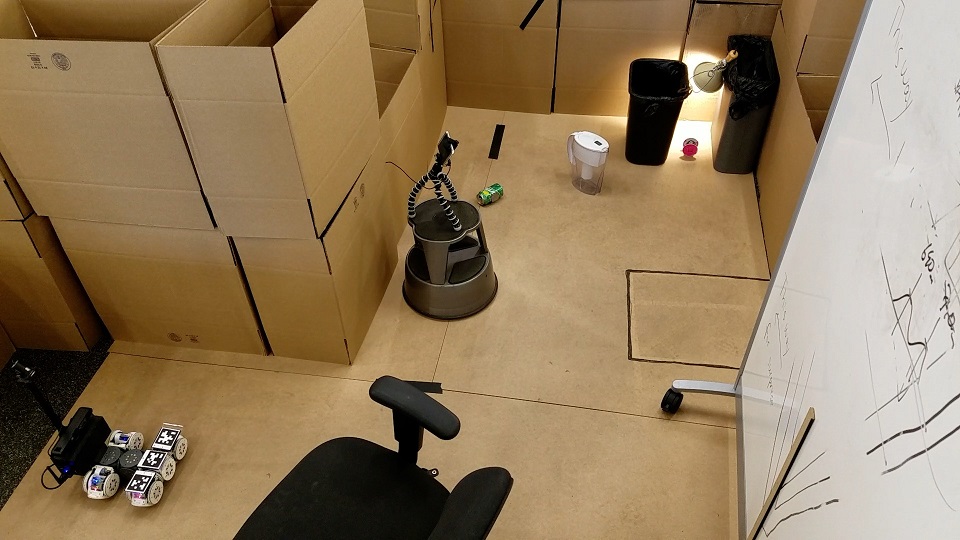}}
  & \pbox{\Rwidth}{\textbf{Demonstration I:} Explore environment to find all pink or green objects and blue dropoff zone. Deliver all objects to dropoff zone.}\\

   & \vspace{-5pt}\\
  \hline
   & \vspace{-5pt}\\
   
  \raisebox{\Rboxheight}{\includegraphics[width=0.3\columnwidth]{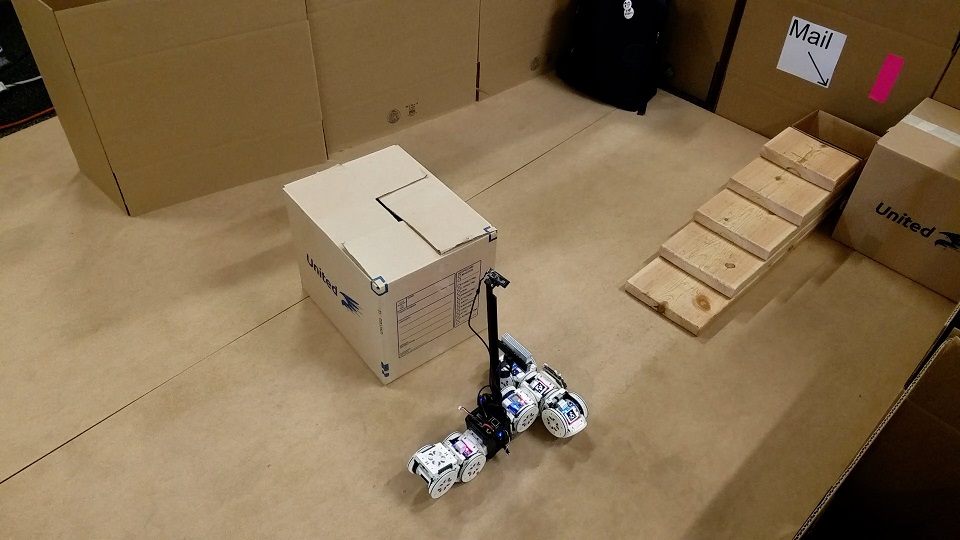}}
  & \pbox{\Rwidth}{\textbf{Demonstration II:} Explore environment to find mailbox, then deliver a circuit to the box.}\\

   & \vspace{-5pt}\\
  \hline
   & \vspace{-5pt}\\
   
  \raisebox{\Rboxheight}{\includegraphics[width=0.3\columnwidth]{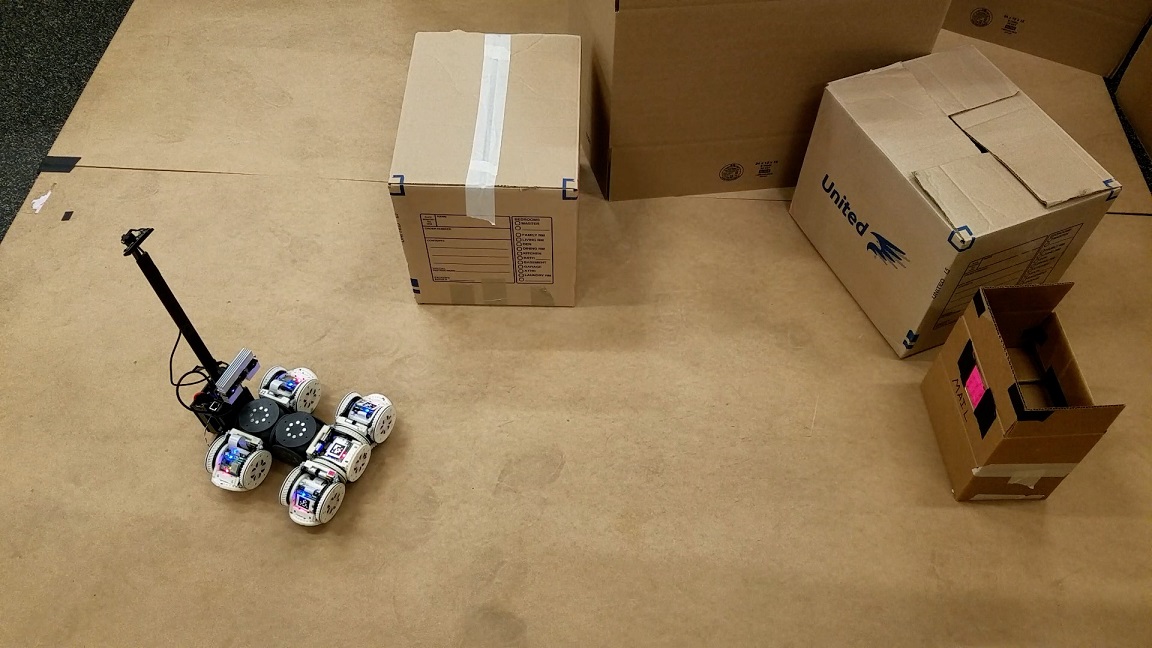}}
  & \pbox{\Rwidth}{\textbf{Demonstration III:} Explore environment to find package, then place a stamp on the package.}\\

   & \\
  \hline
  \end{tabular}
  \caption{Environments and tasks for hardware demonstrations}
  \label{table:task-compare}
  \end{subfigure}
  \setlength{\tabcolsep}{6pt} 
\caption{Environments and Tasks for Demonstrations}
\label{fig:envs}
\end{figure}

Figure \ref{fig:envs} shows the environments used for each demonstration, and Figure \ref{fig:experiments} shows snapshots during each of the demonstrations. A video of all three demonstrations is available as part of the supplementary material.

In Demonstration I, the robot must find, retrieve, and deliver all pink- and green-colored metal garbage to a designated drop-off zone for recycling, which is marked with a blue square on the wall. The demonstration environment contains two objects to be retrieved: a green soda can in an unobstructed area, and a pink spool of wire in a narrow gap between two trash cans. Various obstacles are placed in the environment to restrict navigation. When performing the task, the robot first explores using the ``Car'' configuration. Once it locates the pink object, it recognizes the surrounding environment as a ``tunnel'' type, and the high-level planner reactively directs the robot to reconfigure to the ``Proboscis'' configuration, which is then used to reach in between the trash cans and pull the object out in the open. The robot then reconfigures to the ``Car,'' retrieves the object, and delivers it to the drop-off zone which the system had previously seen and marked during exploration. Figure \ref{fig:octomap} shows the resulting three-dimenstional (3D) map created from simultaneous localization and mapping (SLAM) during the demonstration.

For Demonstrations II and III, the high-level task specification is the following: Start with an object, explore until finding a delivery location, and deliver the object there. Each demonstration uses a different environment. For Demonstration II, the robot must place a circuit board in a mailbox (marked with pink-colored tape) at the top of a set of stairs with other obstacles in the environment. For Demonstration III, the robot must place a postage stamp high up on the box that is sitting in the open.

For Demonstration II, the robot begins exploring in the ``Scorpion'' configuration. Shortly, the robot observes and recognizes the mailbox, and characterizes the surrounding environment as ``stairs.'' Based on this characterization, the high-level planner directs the robot to use the ``Snake'' configuration to traverse the stairs. Using the 3D map and characterization of the environment surrounding the mail bin, the robot navigates to a point directly in front of the stairs, faces the bin, and reconfigures to the ``Snake'' configuration. The robot then executes the stair climbing gait to reach the mail bin, and drops the circuit successfully. It then descends the stairs and reconfigures back to the ``Scorpion'' configuration to end the mission.

For Demonstration III, the robot begins in the ``Car'' configuration, and cannot see the package from its starting location.  After a short period of exploration, the robot identifies the pink square marking the package.  The pink square is unobstructed, but is approximately 25cm above the ground; the system correctly characterizes this as the ``high''-type environment, and recognizes that reconfiguration will be needed to reach up and place the stamp on the target.  The robot navigates to a position directly in front of the package, reconfigures to the ``Proboscis'' configuration, and executes the ``highReach'' behavior to place the stamp on the target, completing its task.

It should be noted that all experiments were run using the same software architecture, same SMORES-EP modules, and system described in this paper. The library of behaviors was extended with new entries as system abilities were added, and minor adjustments were made to motor speeds, SLAM parameters, and the low-level reconfiguration controller. In addition, Demonstrations II and III used a newer, improved 3D sensor, and therefore a different sensor driver was used from Demonstration I.
%
\section{Discussion}
\label{sec:discussion}
%
This paper presents the first modular robot system to autonomously complete high-level tasks by reactively reconfiguring in response to its perceived environment and task requirements. In addition, putting the entire system to the test in hardware demonstrations revealed several opportunities for future improvement in such systems. Modular self-reconfigurable robots are by their nature mechanically distributed, and as a result lend themselves naturally to distributed planning, sensing, and control. Most past systems have used entirely distributed frameworks \cite{Yim2007, Rubenstein2004,Murata2006,Dorigo2005,Mondada2005,o2010self}. Our system is designed differently.  It is distributed at the low level (hardware), but centralized at the high level (planning and perception), leveraging the advantages of both design paradigms.

The three scenarios in the demonstrations showcase a range of different ways SMORES-EP can interact with environments and objects: movement over flat ground, fitting into tight spaces, reaching up high, climbing over rough terrain, and manipulating objects.  This broad range of functionality is only accessible to SMORES-EP by reconfiguring between different morphologies.

The high-level planner, environment characterization tools, and library work together to allow tasks to be represented in a flexible and reactive manner. For example, at the high level, Demonstrations II and III are the same task: deliver an object at a point of interest.  However, after characterizing the different environments (``High'' in II, ``Stairs'' in III), the system automatically determines that different configurations and behaviors are required to complete each task:  the Proboscis to reach up high, and the Snake to climb the stairs.  
Similarly, in Demonstration I there is no high-level distinction between the green and pink objects - the robot is simply asked to retrieve all objects it finds.  The sensed environment once again dictates the choice of behavior: the simple problem (object in the open) is solved in a simple way (with the Car configuration), and the more difficult problem (object in tunnel) is solved in a more sophisticated way (by reconfiguring into the Proboscis).
This level of sophistication in control and decision making goes beyond the capabilities demonstrated by past systems with distributed architectures.

Centralized sensing and control during reconfiguration, provided by AprilTags and a centralized path planner, allowed our implementation to transform between configurations more rapidly than previous distributed systems. 
Each reconfiguration action (a module disconnecting, moving, and reattaching) takes about one minute.  In contrast, past systems that utilized distributed sensing and control required 5-15 minutes for single reconfiguration actions \cite{Yim2007, Rubenstein2004,Murata2006}, which would prohibit their use in the complex tasks and environments that our system demonstrated.

Through the hardware demonstrations performed with our system, we observed several challenges and opportunities for future improvement with autonomous perception-informed modular systems. All SMORES-EP body modules are identical, and therefore interchangeable for the purposes of reconfiguration.  However, the sensor module has a substantially different shape than a SMORES-EP body module, which introduces heterogeneity in a way that complicates motion planning and reconfiguration planning.  Configurations and behaviors must be designed to provide the sensor module with an adequate view, and to support its weight and elongated shape.  Centralizing sensing also limits reconfiguration: modules can only drive independently in the vicinity of the sensor module, preventing the robot from operating as multiple disparate clusters. 

Our high-level planner assumes all underlying components are reliable and robust, so failure of a low-level component can cause the high-level planner to behave unexpectedly, and result in failure of the entire task.  Table \ref{table:errors} shows the causes of failure for 24 attempts of Demonstration II (placing the stamp on the package).  
Nearly all failures are due to an error in one of the low-level components the system relies upon, with
42\% of failure due to hardware errors and 38\% due to failures in low-level software (object recognition, navigation, environment characterization).
This kind of cascading failure is a weakness of centralized, hierarchical systems: distributed systems are often designed so that failure of a single unit can be compensated for by other units, and does not result in global failure.

This lack of robustness represents a challenge, but steps can be taken to address it.  Open-loop behaviors (like stair-climbing and reaching up to place the stamp) were vulnerable to small hardware errors and less robust against variations in the environment. For example, if the height of stairs in the actual environment is higher than the property value of the library entry, the stair-climbing behavior is likely to fail. Closing the loop using sensing made exploration and reconfiguration significantly less vulnerable to error.  Future systems could be made more robust by introducing more feedback from low-level components to high-level decisions making processes, and by incorporating existing high-level failure-recovery frameworks \cite{Maniatopoulos16icra}.  Distributed repair strategies could also be explored, to replace malfunctioning modules with nearby working ones on the fly \cite{tomita1999self}.

To implement our perception characterization component, we assumed a simplified set of environment types and implemented a simple characterization function to distinguish between them. This function does not generalize very well to completely unstructured environments and also is not very scalable. Thus, to expand the system to work well for more realistic environments and to distinguish between a large number of environment types, a more general characterization function should be implemented.


\section{Methods and Materials}\label{sec:system}
\begin{figure}[H]
\begin{center}
\includegraphics[width=0.8\columnwidth]{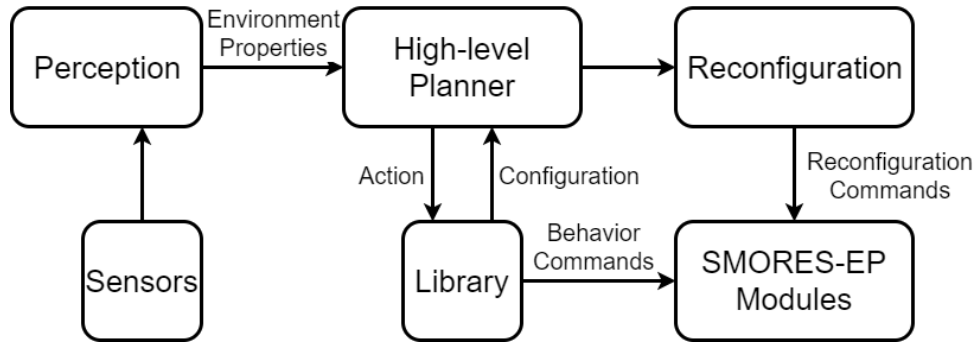}
\caption{System Overview Flowchart}
\label{fig:overview}
\end{center}
\end{figure}

The following sections discuss the role of each component within the general system architecture. Inter-process communication between the many software components in our implementation is provided by the Robot Operating System (ROS)\footnote{http://www.ros.org}. Figure \ref{fig:overview} gives a flowchart of the entire system. For more details of the implementation used in the demonstrations see the Supplementary Materials.

%


\subsection{Hardware} 
\label{sec:hardware}
\paragraph{SMORES-EP Modular Robot:} \label{sec:smores}
Each SMORES-EP module is the size of an 80mm cube
and has four actuated joints, including two wheels that can be
used for differential drive on flat ground \cite{tosun2016design},
\cite{tosun2017paintpots}.  The modules are equipped
with electro-permanent (EP) magnets that allow any face of one module to connect to
any face of another, allowing the robot to self-reconfigure. The magnetic faces
can also be used to attach to objects made of ferromagnetic materials (e.g. steel). 
The EP magnets require very little energy to connect and disconnect, and no energy to maintain their attachment force of 90N \cite{tosun2016design}.

Each module has an onboard battery, microcontroller, and WiFi
module to send and receive messages.  In this work, clusters of SMORES-EP
modules are controlled by a central computer running a Python program that
sends WiFi commands to control the four DoF and magnets of each module.
Wireless networking is provided by a standard off-the-shelf  router, with a range of about 100 feet, and commands to a single module can be received at a rate of about 20hz.
Battery life is about one hour (depending on motor, magnet, and radio usage).

%

\begin{figure}[H]
  \begin{subfigure}{0.5\columnwidth}
  \begin{center}
  \includegraphics[height=2in]{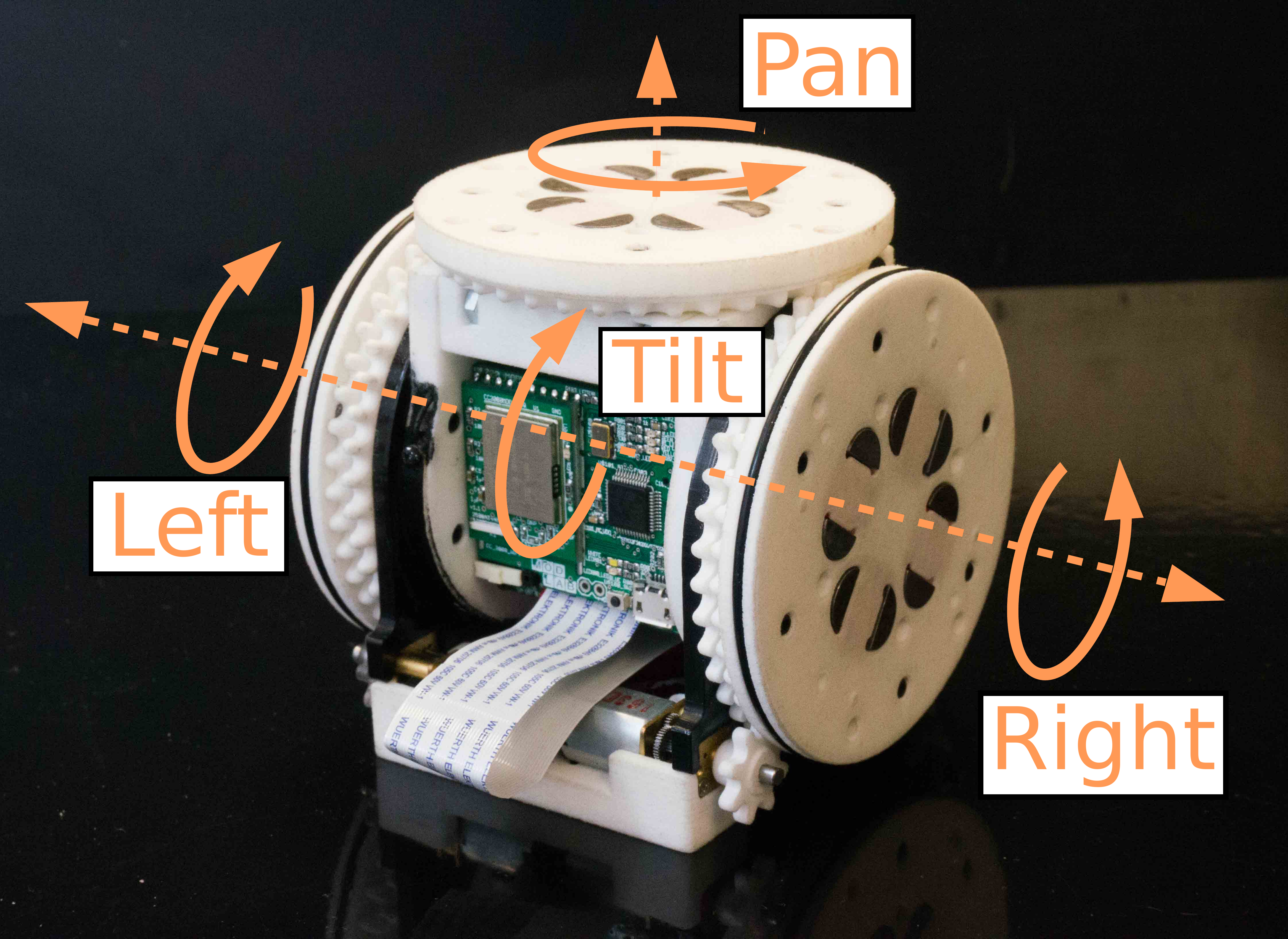}
  \end{center}
  \caption{SMORES-EP module}
  \label{fig:smores-module}
  \end{subfigure}
  \begin{subfigure}{0.5\columnwidth}
  \begin{center}
  \includegraphics[width=0.7\textwidth]{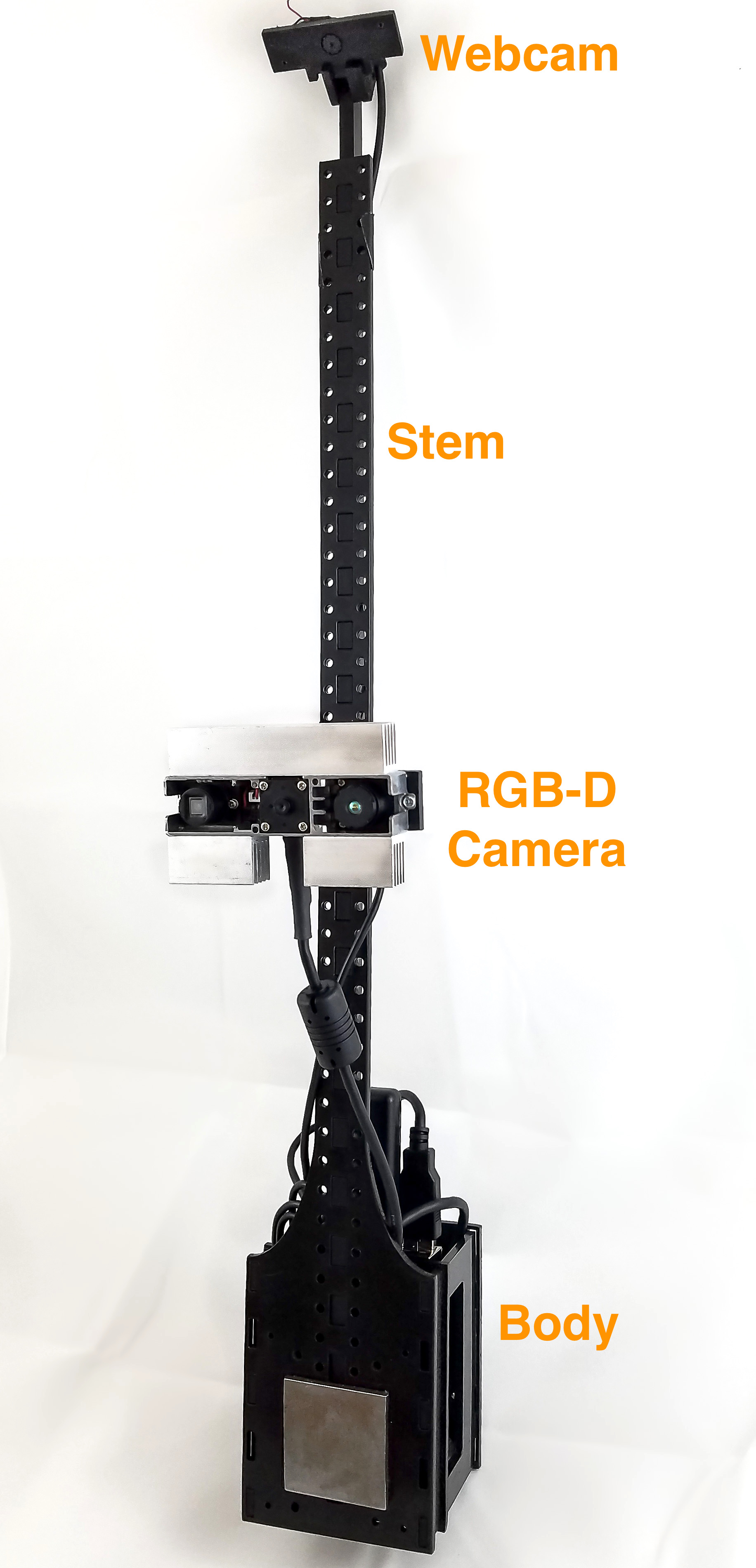}
  \caption{Sensor Module with labelled components.  UP board and battery are inside the body.}
  \label{fig:sensor-module}
  \end{center}
  \end{subfigure}
  \caption{SMORES-EP Module and Sensor Module}
\end{figure}

%
\paragraph{Sensor Module:} 
\label{sec:sensor_module}
SMORES-EP modules have no sensors that allow them to gather information about their environment. To enable autonomous operation, we introduce a \textit{sensor module}, shown in Figure~\ref{fig:sensor-module}.
The sensor module used in our demonstrations was designed to work with SMORES-EP, and is shown in Figure~\ref{fig:sensor-module}.
The body of the sensor module is a 90mm $\times$ 70mm $\times$ 70mm box with thin steel plates on its front and back that allow SMORES-EP modules
to connect to it.
Computation is provided by an UP computing board with an Intel Atom 1.92 GHz
processor, 4 GB memory, and a 64 GB hard drive. A USB WiFi adapter provides
network connectivity. A front-facing Orbecc Astra Mini camera provides RGB-D
data, enabling the robot to explore and map its environment and recognize
objects of interest.  A thin stem extends 40cm above the body, supporting a
downward-facing webcam. This camera provides a view of a  0.75m $\times$ 0.5m area
in front of the sensor module, and is used to track AprilTag
\cite{olson2011apriltag} fiducials for reconfiguration. A 7.4V, 2200mAh LiPo
battery provides about one hour of running time.

A single sensor module carried by the cluster of SMORES-EP modules provides centralized sensing and computation.  Centralizing sensing and computation has the advantage of facilitating control, task-related decision making, and rapid reconfiguration, but has the disadvantage of introducing physical heterogeneity, making it more difficult to design configurations and behaviors.  The shape of the sensor module can be altered by attaching lightweight cubes, which provide passive structure to which modules can connect.  Cubes have the same 80mm form factor as SMORES-EP modules, with magnets on all faces for attachment. 
%

\subsection{Perception and Planning for Information}
\label{sec:exploration}

Completing tasks in unknown environments requires the robot to explore and gain information about its surroundings, and use that information to inform actions and reconfiguration.
Our system architecture includes active perception components to perform SLAM, choose waypoints for exploration, and recognize objects and regions of interest.  It is also includes a framework to characterize the environment in terms of robot capabilities, allowing the high-level planner to reactively reconfigure the robot to adapt to different environment types. Implementations of these tools should be selected to fit the MSRR system being used and types of environments expected to be encountered.

\begin{figure}[H]
\begin{subfigure}[t]{0.45\columnwidth}
    \centering
    \begin{subfigure}[t]{0.45\columnwidth}
        \includegraphics[width=\textwidth]{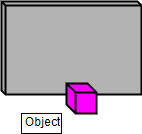}
        \caption{\textbf{``free'}' environment}
    \end{subfigure} \ \ \ \ \ \
    \begin{subfigure}[t]{0.45\columnwidth}
        \includegraphics[width=\textwidth]{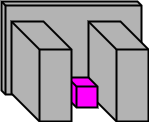}
        \caption{\textbf{``tunnel''} environment}
    \end{subfigure}
    
    \begin{subfigure}[t]{0.45\columnwidth}
        \includegraphics[width=\textwidth]{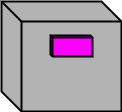}
        \caption{\textbf{``high''} environment}
    \end{subfigure} \ \ \ \ \ \
    \begin{subfigure}[t]{0.45\columnwidth}
        \includegraphics[width=\textwidth]{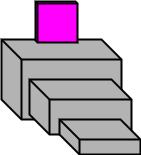}
        \caption{\textbf{``stairs''} environment}
    \end{subfigure}
      \caption{Environment characterization types.}
      \label{fig:characters}
\end{subfigure}
%
\begin{subfigure}{0.5\columnwidth}
\begin{center}
\includegraphics[width=\columnwidth]{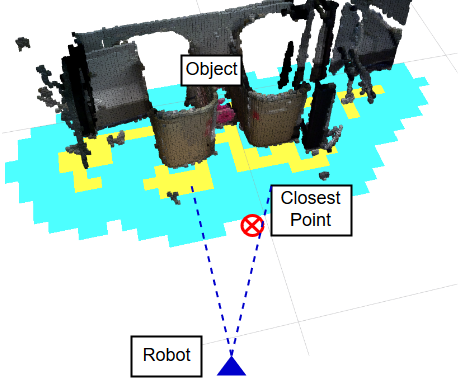}
\caption{An example of a \textbf{tunnel} environment characterization. Yellow grid cells are occupied, light blue cells are unreachable resulting from bloating obstacles.}
\label{fig:characterization}
\end{center}
\end{subfigure}
\caption{Environment Characterization}
\end{figure}

Environment characterization is done using a discrete classifier (using the 3D occupancy grid of the environment as input) to distinguish between a discrete set of environment types corresponding to the library of robot configurations and gaits. To implement our system for a particular MSRR, the classification function must be defined by the user to classify the desired types of environments. For our proof-of-concept hardware demonstrations, we assumed a simplified set of possible environment types around objects of interest. We assumed the object of interest must be in one of four environment types shown in Figure \ref{fig:characters}: ``tunnel" (the object is in a narrow corridor), ``stairs" (the object is at the top of low stairs), ``high" (the object is on a wall above the ground), and ``free" (the object is on the ground with no obstacles around). Our implemented function performs characterization as follows: When the system recognizes an object in the environment, the characterization function evaluates the 3D information in the object's surroundings. It creates an occupancy grid around the object location, and denotes all grid cells within a robot-radius of obstacles as unreachable (illustrated in Figure~\ref{fig:characterization}). The algorithm then selects the closest reachable point to the object within $20^o$ of the robot's line of sight to the object. If the distance from this point to the object is greater than a threshold value and the object is on the ground, the function characterizes the environment as a ``tunnel''. If above the ground, the function characterizes the environment as a ``stairs'' environment. If the closest reachable point is under the threshold value, the system assigns a ``free'' or ``high'' environment characterization, depending on the height of the colored object.

Based on the environment characterization and target location, the function also returns a waypoint for the robot to position itself to perform its task (or to reconfigure, if necessary).  In Demonstration II, the environment characterization algorithm directs the robot to drive to a waypoint at the base of the stairs, which is the best place for the robot to reconfigure and begin climbing the stairs.

Our implementation for other components of the perception architecture use previous work and open-source algorithms. The RGB-D SLAM software package RTAB-MAP\cite{rtabmap} provides mapping and robot pose. The system incrementally builds a 3D map of the environment and stores the map in an efficient octree-based volumetric map using Octomap\cite{octomap}. The Next Best View algorithm by Daudelin et. al.\cite{Daudelin2017} enables the system to explore unknown environments by using the current volumetric map of the environment to estimate the next reachable sensor viewpoint that will observe the largest volume of undiscovered portions of objects (the Next Best View). In the example object delivery task, the system begins the task by iteratively navigating to these Next Best View waypoints to explore objects in the environment until discovering the dropoff zone.

To identify objects of interest in the task (such as the dropoff zone), we implemented our system using color detection and tracking.  The system recognizes colored objects using CMVision\footnote{CMVision: http://www.cs.cmu.edu/$\sim$jbruce/cmvision/}, and tracks them in 3D\footnote{Lucas Coelho Figueiredo: https://github.com/lucascoelho91/ballFollower} using depth information from the onboard RGB-D sensor. Although we implement object recognition by color, more sophisticated methods could be used instead, under the same system architecture.

%



\subsection{Library of Configurations and Behaviors}
\label{sec:supplement-configuration-specifics}

A library-based framework was used to organize user-designed configurations and behaviors for the SMORES-EP robot.
Users can create designs for modular robot using our simulation tool and save designs to a library.
Configurations and behaviors are labeled with properties, which are high-level descriptions of behaviors.
Specifically, environment properties specify the appropriate environment that the behavior is designed for (e.g. a 3 module-high ledge) and behavior properties specify the capabilities of the behavior (e.g. climb). 
Therefore in this framework, a library entry is defined as $l = (C,B_C,P_b,P_e)$ where $C$ is a robot configuration, $B_C$ is the behavior of $C$, $P_b$ is a set of behavior properties describing the capabilities of the behavior, and $P_e$ is a set of environment properties. The high-level planner then can select appropriate configurations and behaviors based on given task specifications and environment information from the perception subsystem to accomplish the robot tasks.
In our Demonstration II, the task specifications require the robot to deliver an object to a mailbox and the environment characterization algorithm reports that the mailbox is in a ``stairs''-type environment.
Then the high-level planner searches the design library for a configuration and behavior that is able\index{self-reconfiguration capabilities} to climb stairs with the object. Each entry is capable of controlling the robot to perform some actions in a specific environment. In Demonstration II, we show a library entry which controls the robot to ``climb'' a ``stairs''-type environment.

To aid users in designing configurations and behaviors, we created a design tool called VSPARC\footnote{\url{www.vsparc.org}}) and made it available online \cite{JingAURO2017}.
Users can use VSPARC to create, simulate and test designs in various environment scenarios with an included physics engine.
Moreover, users can save their designs of configurations (connectivity among modules) and behaviors (joint commands for each module) in plain text files on our server and share them with other users.
All behaviors designed in VSPARC can be used to directly control the SMORES-EP robot system to perform the same action.
Table~\ref{table:1} lists ten entries for four different configurations that are used in this work. 


\subsection{Reconfiguration}
\label{sec:reconfiguration}
When the high-level planner decides to use a new configuration during a task, the robot must reconfigure.
We have implemented tools for mobile reconfiguration with SMORES-EP, taking advantage of the fact that individual modules can drive on flat surfaces.
As discussed in Section~\ref{sec:hardware}, a downward-facing camera on the Sensor Module provides a view of a $0.75\text{m}\times0.5\text{m}$ area on the ground in front of the sensor module.  
Within this area, the localization system provides pose for any module equipped with an AprilTag marker to perform reconfiguration. 
Given an initial configuration and a goal configuration, the reconfiguration controller commands a set of modules to disconnect, move and reconnect to form the new topology of the goal configuration. 
Currently, reconfiguration plans from one configuration to another are created manually and stored in the library. However the framework can work with existing assembly planning algorithms (\cite{Werfel2007,Seo2013}) to generate reconfiguration plans automatically.
Figure~\ref{fig:reconf} shows reconfiguration from the ``Car'' to the ``Proboscis'' during Demonstration 1.
%
\subsection{High-Level Planner}
\label{sec:supplement-high-level}

In our architecture, the high-level planner subsystem provides a framework for users to specify robot tasks using a formal language, and generates a centralized controller that directs robot motion and actions based on environment information.
Our implementation is based on the Linear Temporal Logic MissiOn Planning (LTLMoP) toolkit, which automatically generates robot controllers from user-specified high-level instructions using synthesis \cite{DBLP:conf/iros/FinucaneJK10,DBLP:journals/trob/Kress-GazitFP09}.
In LTLMoP, users describe the desired robot tasks with high-level specifications over a set of Boolean variables, and provide mapping from each variable to a robot sensing or action function.
In our framework, users do not specify the exact configurations and behaviors used to complete tasks, but rather specify constraints and desired outcomes for each Boolean variable using properties from the robot design library.
LTLMoP automatically converts the specification to logic formulas, which are then used to synthesize a robot controller that satisfies the given tasks (if one exists).
The high-level planner determines configurations and behaviors associated to each Boolean variable based on properties specified by users and continually executes the synthesized robot controller to react to the sensed environment.

Consider the robot task in Demonstration II, the user indicates that the robot should \textbf{explore} until it locates the \textbf{mailBox}, then \textbf{drop} the object off.
In addition, the user describes desired robot actions in terms of properties from the library.
The high-level planner then generates a discrete robot controller that satisfies the given specifications.
If no controller can be found or no appropriate library entries can implement the controller, users are advised to change the task specifications or add more behaviors to the design library.

The high-level planner coordinates each component of the system to control our MSRR to achieve complex tasks.
At the system level, the sensing components gather and process environment information for the high-level planner, which then takes actions based on the given robot tasks by invoking appropriate low-level behaviors.
In Demonstration II, when the robot is asked to deliver the object, the perception subsystem informs the robot that the mailbox is in a ``stairs''-type environment.
Therefore, the robot self-reconfigures to a ``Snake'' configuration to climb the stairs and deliver the object.



%
\section*{Acknowledgments}
This work was funded by NSF grant numbers CNS-1329620 and CNS-1329692.


\bibliographystyle{Science}
\bibliography{references}

\section*{Figures and Tables}


\begin{figure}[H]
  \begin{subfigure}{\columnwidth}
  \begin{small}
  \begin{tabular}{c c c}
    \pbox{0.32\textwidth}{
        \includegraphics[width=0.32\textwidth]{images/overhead_starting.jpg}
        1. Demonstration environment} & 
    \pbox{0.32\textwidth}{
        \includegraphics[width=0.32\textwidth]{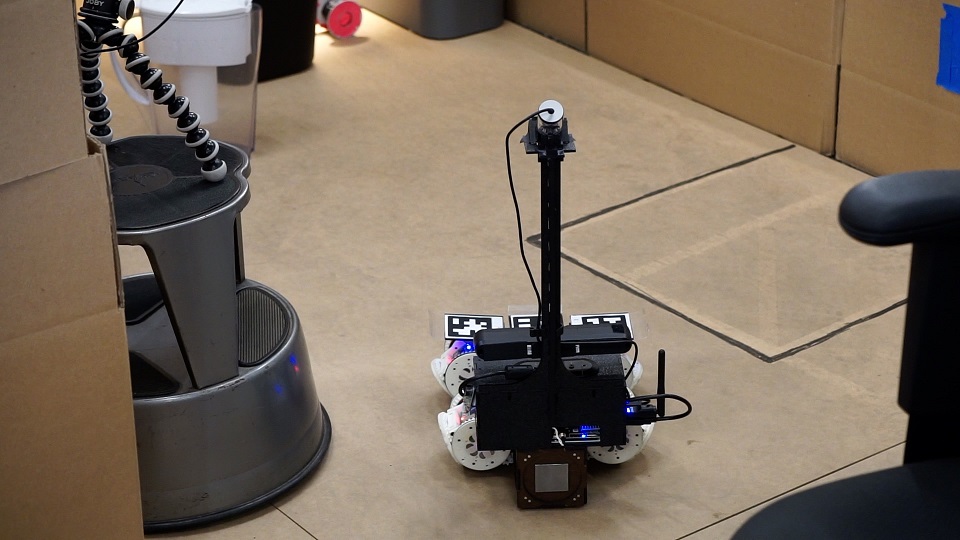}
        2. Exploration of environment} &
    \pbox{0.32\textwidth}{
        \includegraphics[width=0.32\textwidth]{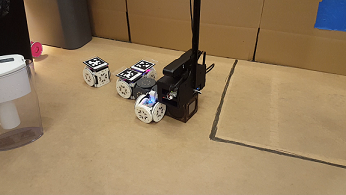}
        3. Reconfiguration}
    \\
    \pbox{0.32\textwidth}{
        \includegraphics[width=0.32\textwidth]{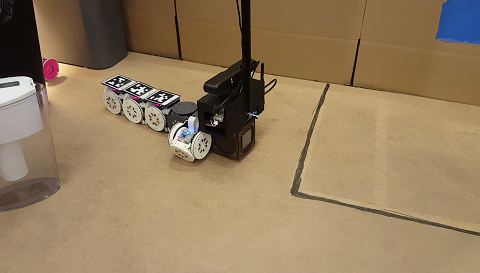}
        4. Retrieving pink object} &
    \pbox{0.32\textwidth}{
        \includegraphics[width=0.32\textwidth]{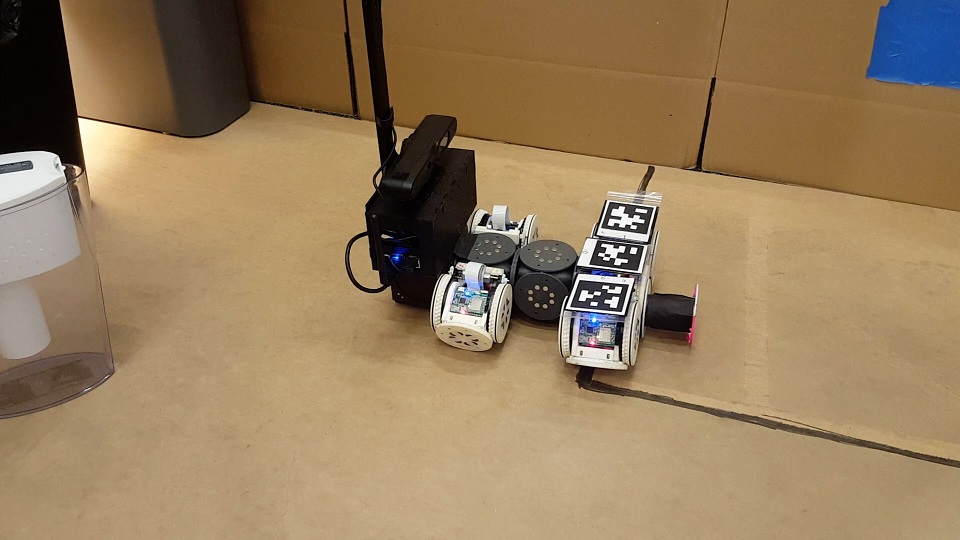}
        5. Delivering an object} &
    \pbox{0.32\textwidth}{
        \includegraphics[width=0.32\textwidth]{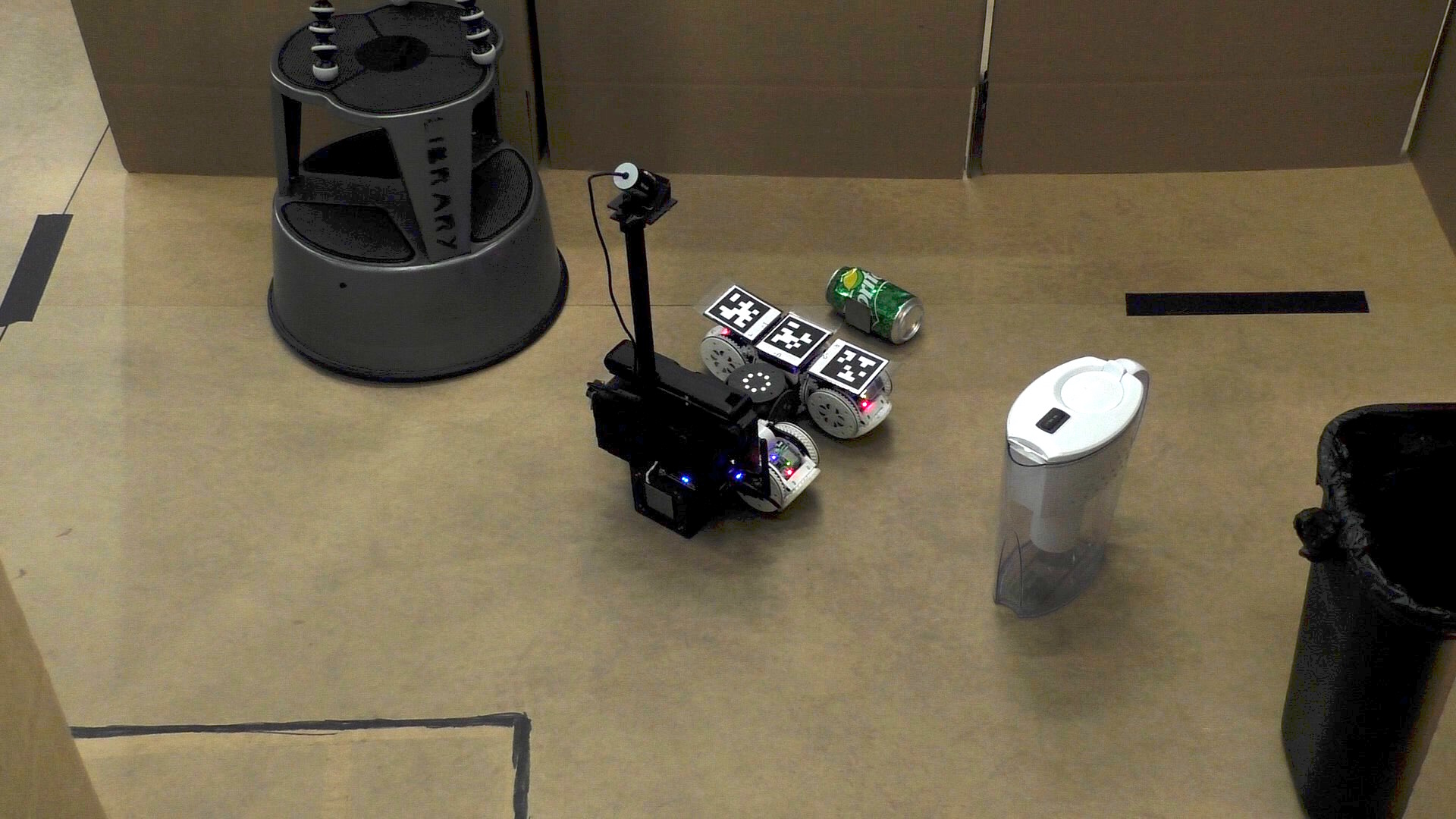}
        6. Retrieving green object}
  \end{tabular}
  \end{small}
        \caption{Phases of Demonstration I.}
  \end{subfigure}
  %
  \begin{subfigure}[t]{\columnwidth}
    \centering
    \begin{small}
    \begin{tabular}{c c}
    \pbox{0.45\textwidth}{
      \includegraphics[width=0.45\textwidth]{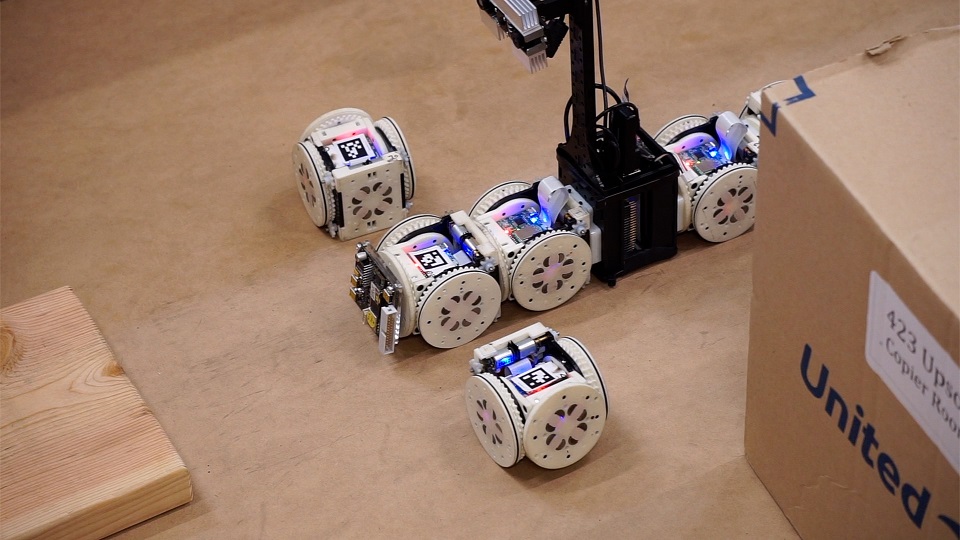}
      1. Reconfiguring to climb stairs} &
    \pbox{0.45\textwidth}{
        \includegraphics[width=0.45\textwidth]{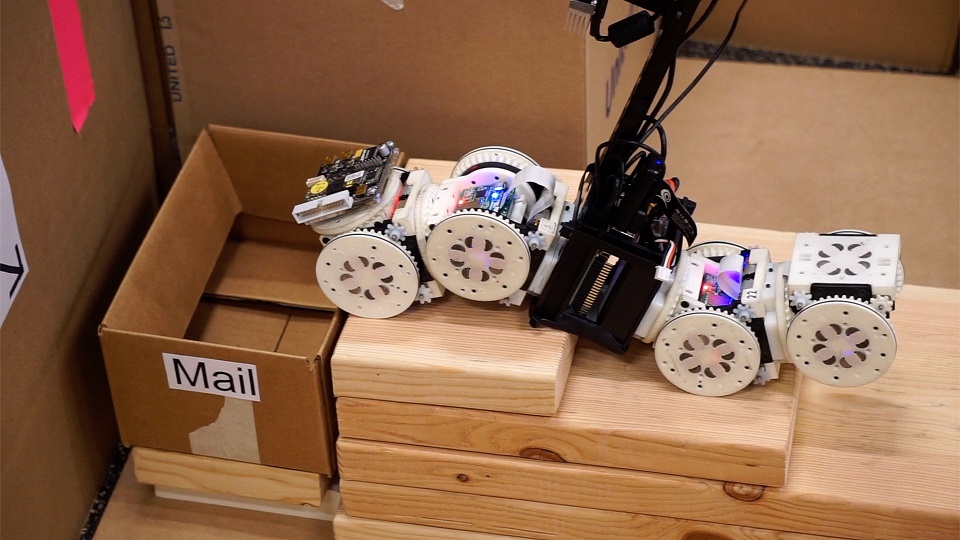}
        2. Successful circuit delivery} 
    \\ 
    \pbox{0.45\textwidth}{
        \includegraphics[width=0.45\textwidth]{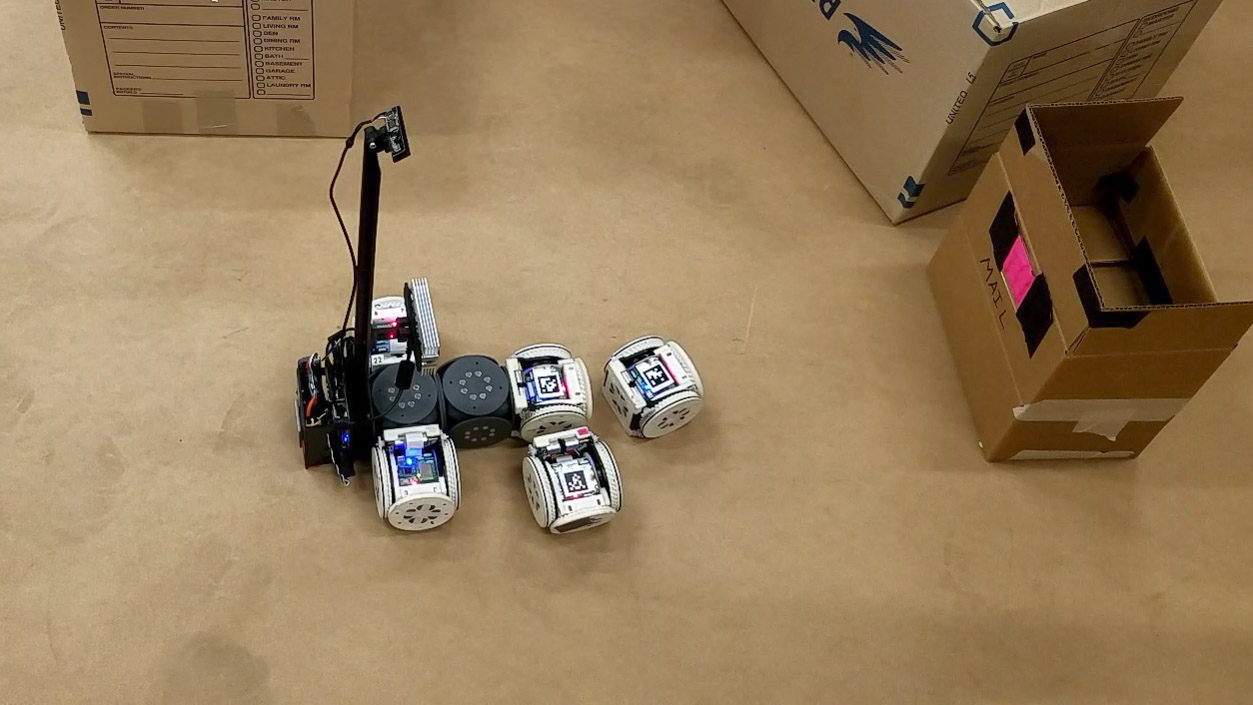}
        1. Reconfiguring to place stamp} &
    \pbox{0.45\textwidth}{
        \includegraphics[width=0.45\textwidth]{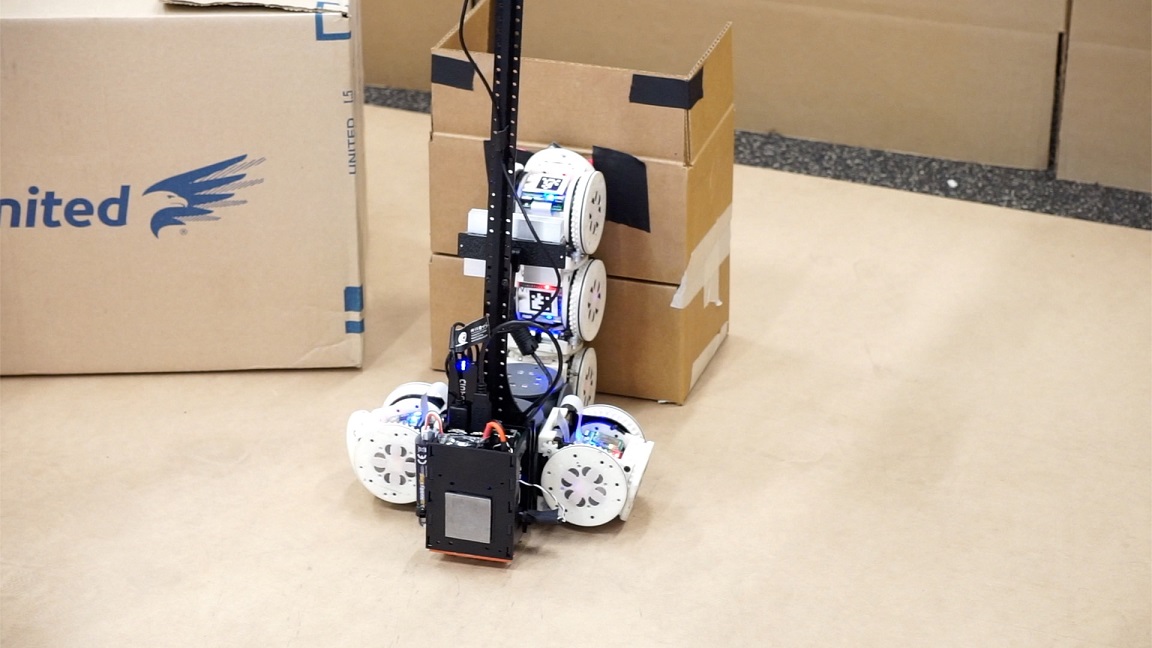}
        2. Successful stamp placement}
    \end{tabular}
    \end{small}
      \caption{Demonstrations II and III.}
  \end{subfigure}  
  \caption{Demonstrations 1, 2, and 3}
  \label{fig:experiments}
\end{figure}

\begin{figure}[H]
\begin{center}
  \includegraphics[width=0.32\columnwidth]{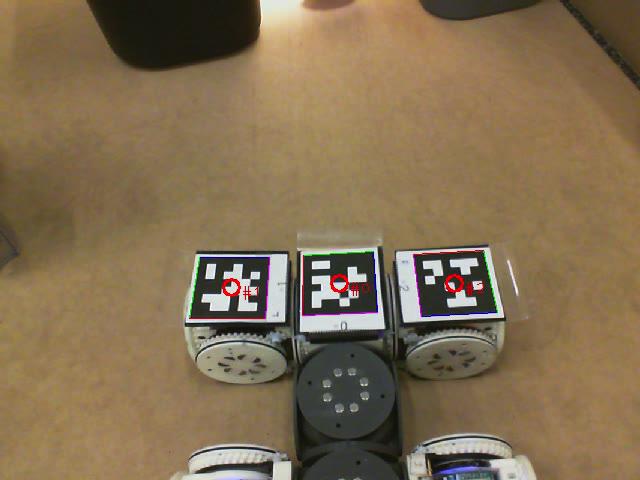}
  \includegraphics[width=0.32\columnwidth]{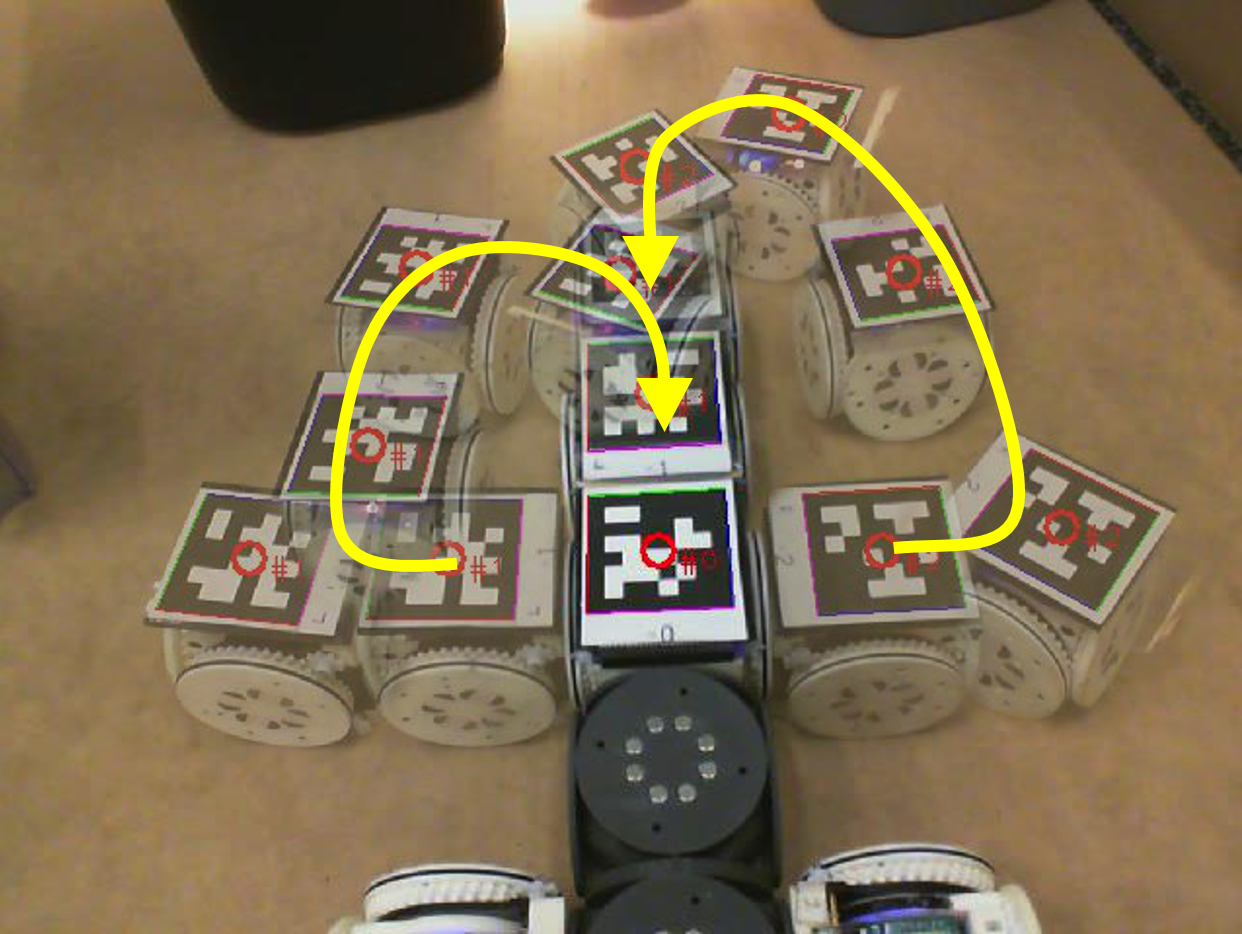}
  \includegraphics[width=0.32\columnwidth]{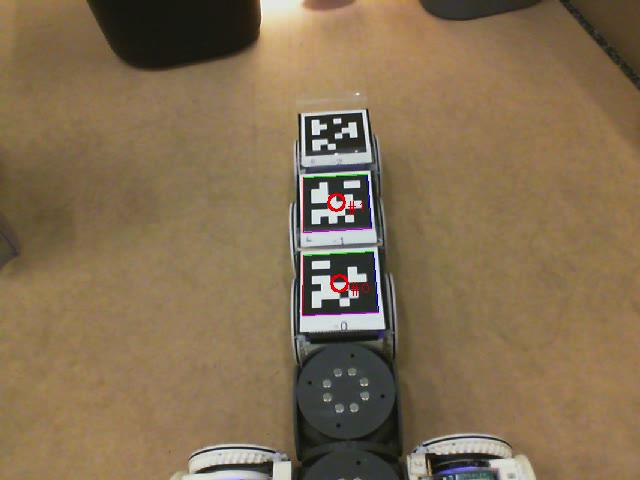}
  \caption{Module movement during reconfiguration. Left: initial configuration (``Car''). Middle: module movement, using AprilTags for localization. Right: final configuration (``Proboscis'').}
  \label{fig:reconf}
\end{center}
\end{figure}

\begin{figure}[H]
\begin{center}
\begin{subfigure}[t]{\columnwidth}
\centering
    \includegraphics[width=\textwidth]{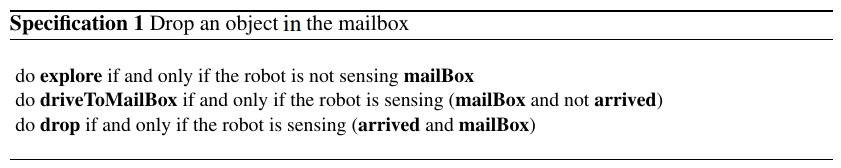}
\caption{Specification for dropping an object in the mailbox.}
\label{fig:spec}
\end{subfigure}

\begin{subfigure}[t]{\columnwidth}
    \centering
    \includegraphics[width=0.4\textwidth]{images/autSimple.png}
\caption{The synthesized controller. A proposition with ``!'' has a value of \lf{}, and \lt{} otherwise.}
\label{fig:autSimple}
\end{subfigure}

\label{fig:specAut}
\caption{A task specification with the synthesized controller.}
\end{center}
\end{figure}



\begin{table}[H]
\centering
\begin{tabular}{ |c|c|c| } 
 \hline
 \multirow{2}{6em}{Configuration} & Behavior & Environment \\
 & properties & Types \\
 \hline
 \multirow{3}{*}{Car} & \textbf{pickUp} & ``free'' \\\cline{2-3}
  & \textbf{drop} & ``free'' \\\cline{2-3}
  & \textbf{drive} & ``free''\\ \hline
 \multirow{3}{*}{Proboscis} & \textbf{pickUp} & ``tunnel'' or ``free''\\ \cline{2-3}
  & \textbf{drop} &``tunnel'' or ``free'' \\ \cline{2-3}
  & \textbf{highReach} & ``high''\\ \hline
 Scorpion & \textbf{drive} & ``free''\\ \hline
 \multirow{3}{*}{Snake} & \textbf{climbUp} & ``stairs''\\ \cline{2-3}
  & \textbf{climbDown} & ``stairs''\\ \cline{2-3}
  & \textbf{drop} & ``stairs'' or ``free''\\
 \hline
\end{tabular}
\caption{A library of robot behaviors}
\label{table:1}
\end{table}

\begin{table}[H]
\centering
\begin{tabular}{|c|c|c|}
\hline
\textbf{Reason of failure} & \textbf{Number of times} & \textbf{Percentage}\\ 
\hline
Hardware Issues & 10 & 41.7\% \\ 
\hline
Navigation Failure & 3 & 12.5\% \\ 
\hline
Perception-Related Errors & 6 & 25\% \\ 
\hline
Network Issues & 1 & 4.2\% \\ 
\hline
Human Error & 4 & 16.7\% \\ 
\hline
\end{tabular}
\caption{Reasons for demonstration failure.}
\label{table:errors}
\end{table}

\pagebreak
\setcounter{page}{1}

\part*{Supplementary Materials}


\section{Additional Commentary on Related Work}\label{sec:related-work}
%
%

Here we provide a more detailed overview of prior work in MSRR systems.  These systems provide partial sets of the capabilities of our system.
 
The Millibot system demonstrated mapping when operating as a swarm. Certain members of the swarm are designated as ``beacons,'' and have known locations. The autonomy of the Millibot swarm is limited: a human operator makes all high-level decisions, and is responsible for navigation using a GUI \cite{Grabowski2000}.

The Swarm-Bots system has been applied in exploration \cite{Dorigo2005} and collective manipulation \cite{Mondada2005} scenarios.  Like the Millibots, some members of the swarm act as ``beacons'' that are assumed to have known location during exploration.  In a collective manipulation task, Swarm-Bots have limited autonomy, with a human operator specifying the location of the manipulation target and the global sequence of manipulation actions.

%
In \cite{o2010self}, Swarm-Bots demonstrate swarm self-assembly to climb a hill.  Robots exhibit phototaxis, with the goal of moving toward a light source.  When robots detect the presence of a hill (using tilt sensors), they aggregate to form a random connected structure to collectively surmount the hill. A similar strategy is employed to cross holes in the ground.  In each case, the swarm of robots is loaded with a single self-assembly controller specific to an \textit{a priori} known obstacle type (hill or hole).  The robots do not self-reconfigure between specific morphologies, but rather self-assemble, beginning as a disconnected swarm and coming together to form a random connected structure.  In our work, a modular robot completes high-level tasks by autonomously self-reconfiguring between specific morphologies with different capabilities.  Our system differentiates between several types of environments using RGB-D data, and may choose to use different morphologies to solve a given high-level task in different environments.    
 
The swarmanoid project (successor to the swarm-bots), uses a heterogeneous swarm of ground and flying robots (called ``hand-'', ``foot-'', and ``eye-'' bots) to perform exploration and object retrieval tasks  \cite{Dorigo2013}. Robotic elements of the swarmanoid system connect and disconnect to complete the task, but the decision to take this action is not made autonomously by the robot in response to sensed environment conditions. While the location of the object to be retrieved is unknown, the method for retrieval is known and constant.

Self-reconfiguration has been demonstrated with several other modular robot systems. CKbot, Conro, and MTRAN have all demonstrated the ability to join disconnected clusters of modules together \cite{Yim2007, Rubenstein2004,Murata2006}. In order to align, Conro uses infra-red sensors on the docking faces of the modules, while CKBot and MTRAN use a separate sensor module on each cluster.  In all cases, individual clusters locate and servo towards each other until they are close enough to dock. These experiments do not include any planning or sequencing of multiple reconfiguration actions in order to create a goal structure appropriate for a task.  Additionally,  modules are not individually mobile, and mobile clusters of modules are limited to slow crawling gaits.  Consequently, reconfiguration is very time consuming, with a single connection requiring 5-15 minutes.

Other work has focused on reconfiguration planning.  Paulos et al. present a system in which self-reconfigurable modular boats self-assemble into prescribed floating structures, such as a bridge \cite{Paulos2015}.  Individual boat modules are able to move about the pool, allowing for rapid reconfiguration.  In these experiments, the environment is known and external localization is provided by an overhead AprilTag system. 

MSRR systems have demonstrated the ability to accomplish low-level tasks such as various modes of locomotion \cite{Yim1994}.
Recent work includes a system which integrates many low-level capabilities of a MSRR system in a design library, and accomplishes high-level user-specified tasks by synthesizing elements of the library into a reactive state-machine \cite{Jing2016}. This system demonstrates autonomy with respect to task-related decision making, but is designed to operate in a fully known environment with external sensing.

Our system goes beyond existing work by using the self-reconfiguration capabilities of an MSRR system to take autonomy a step further.  The system uses perception of the environment to inform the choice of robot configuration, allowing the robot to adapt its abilities to surmount challenges arising from \textit{a priori} unknown features in the environment. Through hardware demonstrations, we show that autonomous self-reconfiguration allows our system to adapt to the environment to complete complex tasks.

%
\section{Library of Configurations and Behaviors}
\label{sec:configuration-specifics-supplement}
In this work, we use the architecture introduced in \cite{Jing2016}. We encode the full set of capabilities of the modular robot, such as driving and picking up items, in a library of robot configurations and behaviors.
To create robot configurations and behaviors, users can utilize our simulator toolbox VSPARC (Verification, Simulation, Programming And Robot Construction \footnote{\url{www.vsparc.org}}) presented in \cite{Jing2016}.
VSPARC allows users to design, simulate and test configurations and behaviors for the SMORES-EP robot system.

Our implementation relies on a framework first presented in \cite{Jing2016}, which is summarized here.
A library entry is defined as $l = (C,B_C,P_b,P_e)$ where:
\begin{itemize}
\item $C$ is the robot \emph{configuration}, specified by the number of modules and the connected structure of the modules.
\item $B_C$ is a \emph{behavior} that $C$ can perform. A behavior is a controller that specifies commands for robot joints to perform a specific movement. 
\item $P_b$ is a set of \emph{behavior properties} that describes what $B_C$ does. 
\item $P_e$ is a set of \emph{environment types} that describe the environments in which this library entry is suitable. 
\end{itemize} 
To specify tasks at the high level, behavior properties $P_b$ are used to describe desired robot actions without explicitly specifying a configuration or behavior.
Environment types $P_e$ specify the conditions under which a behavior can be used.
This allows the high-level planner to match environment characterizations from the perception subsystem with configurations and behaviors that can perform the task in the current environment. 
In Demonstration II, when the environment characterization algorithm reports that the mailbox is located in a ``stairs''-type environment, the high-level planner queries the library for configurations that can climb stairs.  
Since the library indicates that current configuration is only capable of driving on flat ground, the high-level planner opts to reconfigure to the stair-climber configuration, and executes its \textbf{climbUp} behavior.

In \cite{Jing2016}, all robot behaviors are \textit{static} behaviors.
That is, once users create a behavior in VSPARC, joint values for each module are fixed and cannot be modified during behavior execution.
Static behaviors, such as a car with a fixed turning radius, do not provide enough maneuverability for the robot to navigate around unknown environments.
In this work, we expand the type of behaviors in the library by using \textit{parametric} behaviors, which were first introduced in \cite{JingAURO2017}.
Parametric behaviors have joint commands that can be altered during run-time, and therefore allow a wider range of motions.
For example, a parametric behavior for a car configuration can be a driving action with two parameters: turning angle and driving velocity.  
The system associates a parametric behavior with a program that generates values of joint commands based on environment information and current robot tasks.
Based on the sensed environment, the perception and exploration subsystem (Section~\ref{sec:exploration}) can generate a collision-free path, which is used to calculate real-time velocity for the robot.
The system then converts the robot velocity to joint values in parametric behaviors at run-time.

To provide an illustrative example, this paper discusses two configurations and their capabilities in detail.
The ``Car'' configuration shown in Figure~\ref{fig:experiments}a-5 is capable of picking
up and dropping objects in a ``free'' environment. In addition, the ``Car'' configuration can locomote on flat terrain. It uses a parametric differential drive behavior to convert a desired velocity vector into motor commands (\textbf{drive} in Table \ref{table:1}).

The ``Proboscis'' configuration shown in Figure~\ref{fig:experiments}a-4 has
a long arm in front, and is suitable for reaching between obstacles in a narrow ``tunnel'' environment to grasp objects or reaching up in a ``high'' environment to drop items.
However, the locomotion behaviors available for this configuration are limited to forward/backward motion, making it unsuitable for general navigation.

This library-based framework allows users to express desired robot actions in an abstract way by specifying behavior properties. For example, if a task specifies that the robot should execute a behavior with the \textbf{drop} property, the system could choose to use either the Car or Proboscis configurations to perform the action, since both have behaviors with the \textbf{drop} property.
The decision of which configuration to use is made during task execution, based on the sensed environment.
For example, if the perception system reports that the environment is of type ``tunnel'', the Proboscis configuration will be used, because the library indicates that it can be used in ``tunnel''-type environments while the Car cannot.

\section{High-Level Planner}
\label{sec:high-level-supplement}

In order to generate controllers from high-level task specifications, we first abstract the robot and environment status as a set of Boolean propositions.
In Demonstration II, the robot action \textbf{drop} is \lt{} if the robot is currently dropping an object in the mailbox (and \lf{} otherwise) and the environment proposition \textbf{mailBox} is \lt{} if the robot is currently sensing a mailbox (and \lf{} otherwise).
Moreover the proposition \textbf{explore} encodes whether or not the robot is currently searching for the target, the mailbox in this case.

By using a library of robot configurations and behaviors as well as environment characterization tools, we can map these high-level abstraction to low-level sensing programs and robot controllers.
As discussed in Section~\ref{sec:configuration-specifics-supplement}, the user specifies high-level robot actions in terms of behavior properties from the library. 
In Demonstration II, our system can choose to do a drop action by executing any behavior from the library which has the behavior property \textbf{drop}, and which also satisfies the current ``stairs''-type environment. If the current robot configuration cannot execute an appropriate behavior, the robot will reconfigure to a different configuration that can.  In this way, the system autonomously chooses to implement  \textbf{drop}  appropriately in response to the sensed environment.
Our system evaluates propositions related to the state of the environment using perception and environment characterization tools in Section~\ref{sec:exploration}. For example, users can map the proposition \textbf{mailBox} to the color tracking function in our perception subsystem, which assigns the value \lt{} to \textbf{mailBox} if and only if the robot is currently seeing a mailbox with the onboard camera.
The system treats propositions, such as \textbf{explore}, that require the robot to navigate in the workspace differently from the other simple robot actions, such as \textbf{drop}.
In this example, users can map \textbf{explore} to behavior property \textbf{drive}, which represents a set of parametric behaviors as discussed in Section~\ref{sec:configuration-specifics-supplement}.
In order to obtain joint values for behaviors at run-time, a path planner in the perception and planning subsystem (Section~\ref{sec:exploration}) takes into account the robot goal as well as the current environment information from the perception subsystem, and generates a collision-free path for the robot to follow.
Our system then converts this path to joint values, which are used to execute the \textbf{drive} behaviors.

Our implementation employs the Linear Temporal Logic MissiOn Planning (LTLMoP) toolkit to automatically generate robot controllers from user-specified high-level instructions using synthesis \cite{DBLP:conf/iros/FinucaneJK10,DBLP:journals/trob/Kress-GazitFP09}.
The user describes the desired robot tasks with high-level specifications over the set of abstracted robot and environment propositions that are mapped to behavior properties from the library.
LTLMoP automatically converts the specification to logic formulas, which are then used to synthesize a robot controller that satisfies the given tasks (if one exists).
The controller is in the form of a finite state automaton, as shown in Figure~\ref{fig:autSimple}.
Each state specifies a set of high-level robot actions that need to be performed, and transitions between states include a set of environment propositions.
Note some  propositions are omitted in Figure~\ref{fig:autSimple} for clarity.
Execution of the high-level controller begins at the predefined initial state in the finite state automaton. In each iteration, LTLMoP determines the values of all environment propositions by calling the corresponding sensing program. Then, LTLMoP chooses the next state in the finite state machine by taking the transition that matches the current value of all environment propositions. 
In the next state, for each robot proposition LTLMoP chooses a behavior from the design library which satisfies both the behavior properties and current environment type.
For example, in Figure~\ref{fig:autSimple} we start in the top state and execute the \textbf{explore} program.
If  the robot senses a mailbox, the value of \textbf{mailBox} becomes \lt{} and therefore the next state is the bottom right state. We then stop the \textbf{explore} program and execute the \textbf{driveToMailBox} program.
We introduce additional constraints to the original task specifications to guarantee that there exist behaviors in the library to implement the synthesized controller.
Since self-reconfiguration is time-consuming, the controller chooses to execute the selected behavior using the current robot configuration whenever possible.
If the current configuration cannot execute the behavior, the controller instructs the robot to reconfigure to one that can, and if multiple appropriate configurations are available, the controller selects one at random.

\section{Reconfiguration}
\label{sec:reconfiguration-supplement}
When the high-level planner decides to use a new configuration during a task, the robot must reconfigure. Our system architecture allows any method for reconfiguration, provided that the method requires no external sensing. SMORES-EP is capable of all three classes of modular self-reconfiguration (chain, lattice, and mobile reconfiguration) \cite{tosun2016design}.  We have implemented tools for mobile reconfiguration with SMORES-EP, taking advantage of the fact that individual modules can drive on flat surfaces as described in Section \ref{sec:hardware}.

Determining the relative positions of modules during mobile self-reconfiguration is an important challenge. 
In this work, the localization method is centralized, using a camera carried by the robot to track AprilTag fiducials mounted to individual modules.
As discussed in Section~\ref{sec:hardware}, the camera provides a view of a $0.75\text{m}\times0.5\text{m}$ area on the ground in front of the sensor module.  
Within this area, the localization system provides pose for any module equipped with an AprilTag marker to perform reconfiguration. 

Given an initial configuration and a goal configuration, the reconfiguration controller commands a set of modules to disconnect, move and reconnect in order to form the new topology of the goal configuration. 
The robot first takes actions to establish the conditions needed for reconfiguration by confirming that the reconfiguration zone is a flat surface free of obstacles (other than the modules themselves).
The robot then sets its joint angles so that all modules that need to detach have both of their wheels on the ground, ready to drive.
Then the robot performs operations to change the topology of the cluster by detaching a module from the cluster, driving, and re-attaching at its new location in the goal configuration, as shown in Figure~\ref{fig:reconf}.
Currently, reconfiguration plans from one configuration to another are created manually and stored in the library. However the framework can work with existing assembly planning algorithms (\cite{Werfel2007,Seo2013}) to generate reconfiguration plans automatically.
Because the reconfiguration zone is free of obstacles, the controller computes collision-free paths offline and stores them as part of the reconfiguration plan.
Once all module movement operations have completed and the goal topology is formed, the robot sets its joints to appropriate angles for the goal configuration to continue performing desired behaviors.

We developed several techniques to ensure reliable connection and disconnection during reconfiguration.  
When a module disconnects from the cluster, the electro-permanent magnets on the connected faces are turned off.  To guarantee a clean break of the magnetic connection, the disconnecting module bends its tilt joint up and down, mechanically separating itself from the cluster. During docking, accurate alignment is crucial to the strength of the magnetic connection \cite{tosun2016design}.  For this reason, rather than driving directly to its final docking location, a module instead drives to a pre-docking waypoint directly in front of its docking location.  At the waypoint, the module spins in place slowly until its heading is aligned with the dock point, and then drives in straight to attach. To guarantee a good connection, the module intentionally overdrives its dock point, pushing itself into the cluster while firing its magnets.
%

\end{document}